# OPTIMAL PARAMETER SELECTION FOR UNSUPERVISED NEURAL NETWORK USING GENETIC ALGORITHM

Suneetha Chittineni[1] and Raveendra Babu Bhogapathi[2]

[1]R.V.R. & J.C. College of Engineering, Guntur, India.
[2]VNR Vignana Jyothi Institute of Engineering & Technology, Hyderabad, India.

## ABSTRACT

*K-means Fast Learning Artificial Neural Network (K-FLANN) is an unsupervised neural network requires two parameters: tolerance and vigilance. Best Clustering results are feasible only by finest parameters specified to the neural network. Selecting optimal values for these parameters is a major problem. To solve this issue, Genetic Algorithm (GA) is used to determine optimal parameters of K-FLANN for finding groups in multidimensional data. K-FLANN is a simple topological network, in which output nodes grows dynamically during the clustering process on receiving input patterns. Original K-FLANN is enhanced to select winner unit out of the matched nodes so that stable clusters are formed with in a less number of epochs. The experimental results show that the GA is efficient in finding optimal values of parameters from the large search space and is tested using artificial and synthetic data sets.*

## 1. INTRODUCTION

Clustering is the task of grouping similar objects in to the same class, while placing dissimilar objects in to different class. All objects in a cluster share the common characteristics. The aim of clustering algorithms is to minimize intra-cluster distance and to maximize inter-cluster distance. The applications of clustering are, similarity searching in medical image data bases, data mining, document retrieval and pattern classification [1]

Artificial neural networks are massively parallel computing systems consisting of an extremely large number of interconnected simple processing elements called neurons [1]. The main features include mapping input patterns to their associated outputs, fault tolerant, and ability to predict new patterns on which they have not been trained [3][4]. Because of this, artificial neural network plays an important role to classify data. The values of initial parameters used by neural network are also essential to obtain best solution. According to Maulik and S Bandyopadhyay [6] applying genetic algorithm is considered as appropriate and natural to select parameters of neural network, to obtain more robust, fast and close approximation and also stated by Elisaveta Shopova and Natasha G. Vaklieva-Bancheva when applied for chemical engineering problems[19].

Genetic algorithm is an evolutionary algorithm that uses the principles of evolution and natural selection to solve hard problems. Genetic algorithm applies genetic operators such as mutation and crossover to evolve the solutions in order select the best solution [2].

Genetic Algorithms [22][8] were initially developed by Bremermann in 1958 and is popularized by J.H. Holland [7][9][10]. They are suitable for the problems in which solution space is very large and doing exhaustive search is unfeasible. Each individual solution is represented using a suitable data structure, and returns one best fitting solution among the set of solutions [7]. The quality of clustering is improved by genetic operators on individual for a fixed number of iterations. Genetic algorithms give better results than the traditional heuristic algorithms because they will never stuck into local maxima. Genetic algorithm works on number of solutions simultaneously. For each solution in the population, the neural network settles in a stable state i.e. in successive iteration there is no change in the formation of clusters. Best solution is selected based on the error rate and the ratio of within the group variance to between the groups variance. By minimizing the two objectives, optimal parameters of neural network are selected.

There are many FLANN models proposed in the literature. The FLANN model was first proposed by Tay and Evans [16], a generalized modification of the Adaptive Resonance Theory (ART 1) by Grossberg and Carpenter [11]. Later Tay and Evans in 1994 developed FLANN II based on the model of Kohonen LVQ network and nearest neighbor was computed using Euclidean distance [12],[16]. Next FLANN II was modified to exploit for continuous input applications [13]. Next K-FLANN was designed from FLANN by including K-means calculations to obtain consistent centroids [14]. But the limitations of K-FLANN are, cluster centroid calculation after each new pattern assigned to the cluster. So 'N' computations are needed for a total of 'N' data patterns. Secondly, Stable clusters are not formed due to the dependency on the data presented sequence [DPS]. The later improvement on K-FLANN includes data point reshuffling which resolves data sequence sensitivity that creates stable clusters [15]. Alex L.P.Tay and Xiang in 2004 proposed GA-KFLANN to determine the parameters tolerance and vigilance [17] and the results are analyzed for a single run of the genetic algorithm.

By referring the previous work [17], the work is based on the following: performance of the GA is verified with a minimum of 60 generations to a maximum of 100 generations. To produce offspring the best parents are selected by Roulette wheel selection. The fitness of chromosomes is computed with CS measure and error rate. Finally, elitism is used to determine the best chromosomes for next iteration.

The main objectives of this investigation are:

1) To find best possible values of neural network parameters.
2) To minimize classification error rate (%) with optimal values of vigilance and tolerance.

## 2 K-MEANS FAST LEARNING ARTIFICIAL NEURAL NETWORK

### 2.1 Architecture

The architecture of the KFLANN is shown in figure 1. K-FLANN is build from ART1 and Kohonen SOM model [11] [18]. It has single input layer. The output layer grows dynamically as new groups are formed in the clustering process. The weights on the arcs give the information about the associated output node for the corresponding input vector. The dynamic formation of output nodes yields K-FLANN as a self-organized clustering ANN model.

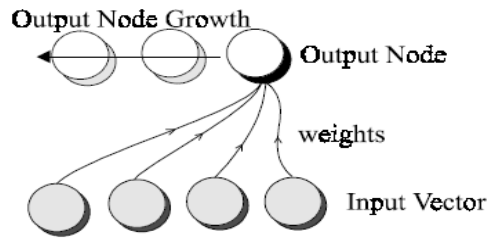

Figure 1 KFLANN architecture

## 2.2 Neural Network Parameters

The two main neural network parameters vigilance and tolerance controls the operation of K-means fast learning artificial neural network (K-FLANN). If there are 'd' features, the set of possible values for vigilance are [1/d, 2/d, … ,d/d]. If it is (1/d), only one feature among the 'd' features must be matched to place the data element in a cluster. The step 3 of the K-FLANN algorithm shows the matching criterion for a pattern. The tolerances of features show the localized variation affecting the individual input features where as the vigilance parameter shows the global variation in the input features. The tolerance setting is defined as the maximum allowable range of specified feature fluctuation. It allows control over the effects of random noise.

## 2.3 Original K-FLANN Algorithm

Step 1  Initialize the network parameters.

Step 2  Present the pattern to the input layer. If there is no output node, Go to Step 6

Step 3  Determine all possible matches of output node using eq. ( 1)

$$\frac{\sum\limits_{i=1}^{d} D\left[\delta_i^2 - \left(w_{ij} - x_i\right)^2\right]}{n}, \qquad D[a]= \begin{cases} 1, a > 0 \\ 0, a \le 0 \end{cases} \quad (1)$$

Step 4  Determine the winner from all matched output nodes using eq. (2)

$$\text{Winner} = \min\left[\sum\limits_{i=1}^{d}\left(w_{ij} - x_i\right)^2\right] \qquad (2)$$

Step 5  Match node is found. Assign the pattern to the match output node. Go to Step 2

Step 6  Create a new output node. Perform direct mapping of the input vector into weight vectors.
    Go to Step 2

Step 7  After completing a single epoch, compute clusters centroid. If centroid points of all clusters unchanged, terminate.
    Else
    Go to Step 8.

Step 8  Find closest pattern to the centroid and re-shuffle it to the top of the data set list, Go to Step 2.

Note:   $\rho$  is the Vigilance Value, $\delta_i$ is the tolerance for ith feature of the input space, $W_{ij}$  used to denote the weight connection of $j_{th}$ output to the ith input (centroid),  $X_i$  represents the $i_{th}$ feature.

### 2.4  Enhanced K-FLANN (EK-FLANN)

The modification in the K-FLANN algorithm is in the step (step 4) of computing best matching unit to form consistent clusters.

**K-FLANN algorithm in step 4 is modified as follows:**

Step 4  Determine the winner from all matched output nodes using the following criteria:

If same match is found

$$\text{Winner} = \min\left[\sum_{i=1}^{d}\left(w_{ij} - x_i\right)^2\right] \tag{3}$$

Else

$$\text{Winner} = \max\left(\frac{\sum_{i=1}^{d}\left[\delta_i^2 - \left(w_{ij} - x_i\right)^2\right]}{n}\right) \tag{4}$$

# 3 GENETIC ALGORITHM

## 3.1 Generating Initial Population

Initial population consists of set of solutions (chromosomes).  Each chromosome consists of encoded strings for two kinds of genes: one is control gene (ConG) and the other is coefficient gene (CoefG)[17]. Control gene is used to determine vigilance parameter ($\rho$). Coefficient gene is used to determine tolerance ($\delta$) values of each feature. The total length of the chromosome is (2*d) +1, where'd' is total number of features exist in the data set.

## 3.2 Structure of Chromosome

Suppose for a data set of 4 features, the length of the chromosome is 9.  ConG has a length of 4 and CoefG has a length of 4. Cong is a binary gene, where as CoefG consists of real numbers. The last column indicates the value of vigilance parameter.  The Cong of length'd' is generated randomly with a set of 0's or 1's.  The number of 1's indicates the requirement match of corresponding features.  Here for 4 features, if the number of 1's in Cong are 3, then it means that the vigilance value ¾=0.75. The CoefG is the tolerance value of each feature is generated randomly whose value is in the range [mindist, maxdist/2,], where 'maxdist' is the maximum distance among data points and 'mindist' is the minimum distance among data points within the feature.

Each solution vector is represented by

| 1 | 0 | 1 | 1 | 0.1588 | 0.0351 | 0.5684 | 0.0121 | 0.75 |
|---|---|---|---|--------|--------|--------|--------|------|

Figure 2 Structure of Solution vector

## 3.3 Selection

Selection plays a major role in GA. It determines how individuals compete for gene survival. Selection always keeps good solutions by picking out bad solutions. The selection is based on the probability of the given individual [19]. In this work, Roulette wheel selection is used to select best chromosomes of the population for reproduction. Same chromosome may be selected more than once in the process of selection when the wheel spins.

Figure 3 illustrates the principle of roulette wheel selection. In this method, 'n individuals are selected by rotating the wheel 'n' times. An individual chromosome may be selected more than once when the wheel spins.

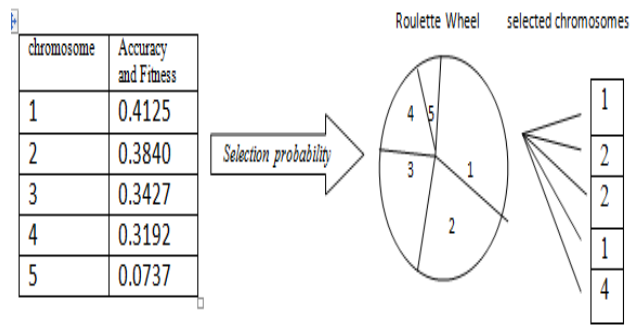

Figure 3  Roulette wheel Selection

**Algorithm for Roulette Wheel Selection**:

Roulette wheel selection can be implemented as follows:

   i.    Find the sum of the fitness of all the population members.
  ii.    Find the probability of all the population members which is equal to the ratio of its fitness and total fitness.
 iii.    Find the cumulative sum of probabilities (CUMPROB).
 iv.    For i =1 :  no. of parents
  v.    Begin
 vi.    Find the difference vector between CUMPROB and random numbers vector.
 vii.    Return the population member whose difference is positive and minimum.
viii.    Go to step 5. Select second parent.
 ix.    i = i+2
  x.    End

### 3.4 Fitness Evaluation

The performance of a clustering algorithm can be evaluated by cluster validity measures such as DB index, Dunn's index…etc. The cluster validity measure CS measure is used for fitness evaluation in GA-KFLANN. Validity indices are used to determine the number of clusters and to find out the best partition. The best partition is known based on the maximum or minimum value of these measures.

### 3.4.1 CS Measure

The cluster validity measure, CS measure was recently proposed by Chou et al. in 2004. For all these measures, the validation is based on the ratio of sum of intra-cluster distances to sum of inter-cluster distances [20].

The cluster representative is computed by taking the mean of data vectors that belong to that cluster. It is shown in eq.5

$$\vec{m}_i = \frac{1}{N_i} \sum_{x_j \in c_i} \vec{x}_j \qquad (5)$$

Assume two data points $\vec{x_i}$ and $\vec{x_j}$. The distance between them is denoted by $d\left(\vec{x_i}, \vec{x_j}\right)$. Then, the CS measure is calculated by eq.6.

$$CS(K) = \frac{\frac{1}{K}\sum_{i=1}^{K}\left[\frac{1}{N_i}\sum_{\vec{x_i} \in c_i} \max_{\vec{X_q} \in C_i}\left\{d\left(\vec{X_i}, \vec{X_q}\right)\right\}\right]}{\frac{1}{K}\sum_{i=1}^{K}\left[\min_{j \in K,\, j \neq i}\left\{d\left(\vec{m_i}, \vec{m_j}\right)\right\}\right]} \qquad (6)$$

$$= \frac{\sum_{i=1}^{K}\left[\frac{1}{N_i}\sum_{\vec{x_i} \in c_i} \max_{\vec{X_q} \in C_i}\left\{d\left(\vec{X_i}, \vec{X_q}\right)\right\}\right]}{\sum_{i=1}^{K}\left[\min_{j \in K,\, j \neq i}\left\{d\left(\vec{m_i}, \vec{m_j}\right)\right\}\right]} \qquad (7)$$

The minimum value of CS index indicates best partition. But the fitness value is proportional to number of clusters. So in the fitness calculation, error rate of a corresponding partition is also considered. The mis-classification rate must be low for the best partition.

Eq. (8) gives the fitness function of a solution vector 'i' based on the result of KFLANN.

$$f\left(\overrightarrow{X_i}\right) = \frac{1}{CS_i(K) * errorrate_i + eps} \tag{8}$$

Where

$CS_i(K)$ - CS measure calculated over K partitions
$errorrate_i$ - mis-classification rate of 'i'
$eps$ - small bias term equal to $2*10^{-6}$ and prevents the denominator of R.H.S from being equal to zero.

## 3.5 Crossover

Crossover operator is used to combine two parent strings of the previous generation to produce better offspring. New offspring is created by exchanging information from the parent strings selected from the mating pool. If the good strings are produced using crossover, then there will be more copies of them in the next mating pool. The generated new offspring may or may not have good characteristics from parent strings, depending up on whether the crossover point is at appropriate place. Uniform crossover is applied to ConG and arithmetic crossover is used for CoefG.

### 3.5.1 Uniform Crossover

Uniform crossover is a simple method to produce offspring by exchanging the features of the parents. The features can be copied either from first parent or second parent based on generated uniform random number. If the number is less than the crossover mask, then the features will be copied from second parent. Otherwise, the features will be copied from first parent. If 'L' is the length of chromosome, then the number of crossing points is 'L/2'

### 3.5.2 Arithmetic Crossover or Convex Crossover

In arithmetic crossover, some arithmetic operation is used to produce new offspring. The parent chromosomes are combined to produce new offspring according to the following equations

$$Offspring1 = a * Parent1 + (1-a) * Parent2 \tag{9}$$

$$Offspring2 = (1-a) * Parent1 + a * Parent2 \tag{10}$$

*Where 'a' is a random weighting factor. The value of 'a' is set to 0.7*

## 3.6 Mutation

Mutation operator is used to invert the bits of the offspring. It helps GA to avoid getting trapped at local maxima. Normal mutation is applied to ConG. Non-uniform mutation is applied to CoefG.

### 3.6.1 Normal Mutation

Mutation probability is set to 0.05. In normal mutation a single bit is selected in a gene when mutation probability less than rand (0, 1).

---

Pseudo code for normal mutation

---

1) Select a gene x at random.
2) If rand (0, 1) < mutation probability then
   Begin
         Select randomly a single bit position 'b' to
          mutate in gene 'x'.
         b'= 1- x (b)
   End
   Else
         x' = x
   End
   End

---

### 3.6.2 Non-uniform Mutation

Non-uniform mutation operator is only used integer and real genes. The value of the selected CoefG gene is replaced based on user defined upper bound and lower bound limits of that gene.

$$x' = x + r\left(u - x\right)\left(1 - \frac{G}{MAXG}\right)^{b} \qquad (11)$$

$$x' = x - r\left(x - l\right)\left(1 - \frac{G}{MAXG}\right)^{b} \qquad (12)$$

Where  b- shape parameter, u,l- upper and lower bounds,  r- Random number chosen uniformly [0, 1], G-current generation number,  MAXG- maximum number of generations

### 3.7 Elitism

Elitism is used to copy best parents to the next generation to replace worst children. When genetic operators are applied on the population, there is a chance of losing the best chromosome. So the best fit chromosome is preserved at each generation. The fitness of offspring is compared with their parent chromosomes.

**The pseudo code for Elitism**:

Begin

1) If fitness (offspring1) is better than fitness (parent1)    then   replace parent1 with offspring1;
   Else if fitness (offspring1) is better than fitness (parent2)   then
   Replace parent2 with offspring1;
   End

2) If fitness (offspring2) is better than fitness (parent2)   then  replace parent2 with offspring2;

Else if fitness (offspring2) is better than fitness (parent1) then
Replace parent1 with offspring1;
End
End

# 4. GENETIC ALGORITHM AND KFLANN

KFLANN works as follows. In the first iteration, the number of output nodes created determines the number of clusters in data set. The number of clusters present in the data set is known from domain knowledge. The number of groups formed for a data set mainly depends on two parameters vigilance and tolerance. The tolerance is computed by the max-min formula [14], [15]. Tolerance tuning is necessary when correct number of clusters is not performed. In such a case, the algorithm runs about 10 to 20 times and consumes more computation time. The search space is very large due to the high tolerance range. The tolerance value of a feature can be chosen in between maximum and minimum distance among the data points of the feature. The range of values for vigilance parameter will increase for a data set with large number of features. It is not possible to perform exhaustive search to select the best values of vigilance and tolerance. So the optimal values of parameters can be determined using genetic algorithm.

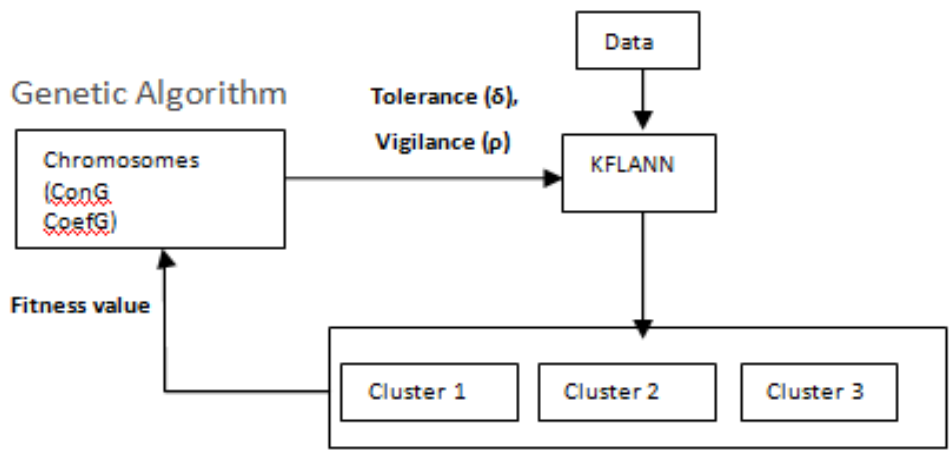

Figure 4 Genetic algorithm applied to KFLANN models

GA consists of set of chromosomes (solutions). Each solution is applied to the K-FLANN to determine the internal structure lies in data set and the algorithm terminates when stable clusters are formed. Figure 4 shows the combination of GA and K-FLANN. The validity measure CS measure is used for fitness calculation. The fitness value is not sufficient to select best fit chromosomes because the fitness value increases as the number of clusters increases. So the best fit chromosomes are selected based on fitness value and error rate as the fitness value is inversely proportional to error rate. For an optimal partition, the error is low and the fitness of chromosome is high. So the error rate is multiplied with fitness function called O*bjectiveFit*. The new population is created by using crossover and mutation. The same process is repeated for 2o generations in each run. The GA is repeated for a maximum of 100 runs.

# 5. EXPERIMENTAL RESULTS

## 5.1 Data Sets Used

The data set is shown in Table 1. It gives the information about the data set name, total number of patterns exists in the data set, number of attributes of a data set, number of classes present in the data set.

Table 1 Description of Data sets used

| S.No. | Data Set | # of patterns | # of features | # of clusters |
|---|---|---|---|---|
| 1 | Iris | 150 | 4 | 3(class 1-50, class 2-50, class 3-50) |
| 2 | Wine | 178 | 13 | 3(class 1-59,class 2-70, class 3-49) |
| 3 | New Thyroid | 215 | 5 | 3(class 1-150,class 2-35, class 3-30) |
| 4 | Haberman | 306 | 3 | 2(class 1-225,class 2-81) |
| 5 | Synthetic data set 1 | 1000 | 2 | 2(class 1-500,class 2-500 ) |
| 6 | Synthetic data set 2 | 1000 | 2 | 2(class 1-500,class 2-500) |
| 7 | Synthetic data set 3 | 1000 | 2 | 2(class 1-500,class 2-500) |
| 8 | Synthetic data set 4 | 500 | 8 | 3(class 1-250, class 2-150, class 3-100 ) |
| 9 | Synthetic data set 5 | 400 | 8 | 3(class 1-150,class 2-150, class 3-100 ) |
| 10 | Synthetic data set 6 | 350 | 8 | 3(class 1-100,class 2-150, class 3-100) |

The artificial data sets used to verify the performance of KFLANN are downloaded from UCI machine learning repository [21]. Synthetic data sets are generated using Matlab functions.

The experiments are conducted without Control Gene (ConG). Control gene is used to set the value for vigilance ($\rho$). For example, for 4 features in the data set, the bits in the control gene are 1 0 1 1, then vigilance value is 0.75. If control gene is considered for clustering, then it is used for feature selection. The features with the bit '1' will take part in clustering process.

## 5.2 Distance Plots

Figures 5 and 6 show the distance plots of artificial data sets and synthetic data sets used in this experiment. The distance plots gives the information about number of clusters exists in a data set. Iris data set has 3 clusters, wine data set has '3' clusters, and so on. Iris data set has three clusters of equal size (1-50,2-50,3-50) . For wine data set, second cluster is large (70/178). For New thyroid data set first cluster is of large size. Iris data set has overlapped clusters (2 and 3)

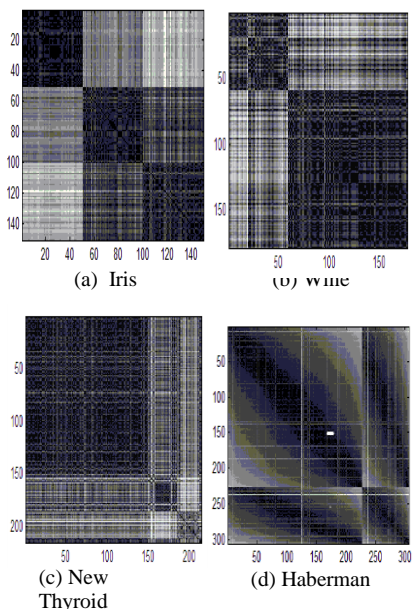

(a) Iris        (b) Wine

(c) New Thyroid        (d) Haberman

Figure 5 Distance plots for artificial data sets

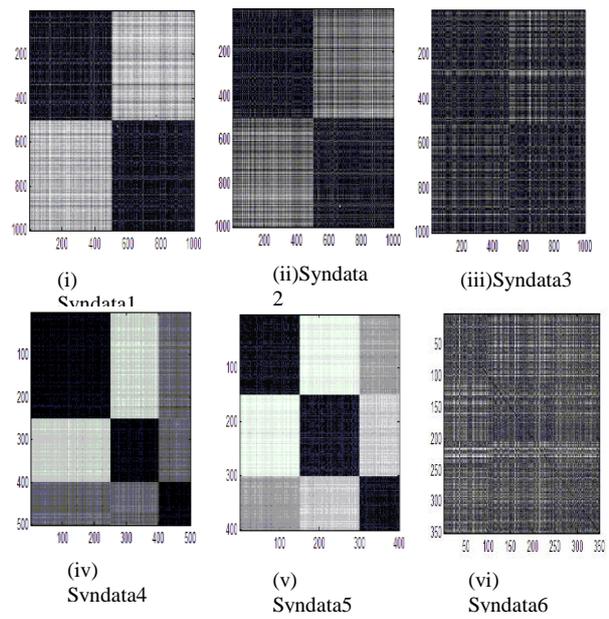

(i) Syndata1        (ii)Syndata2        (iii)Syndata3

(iv) Syndata4        (v) Syndata5        (vi) Syndata6

Figure 6 Distance plots for synthetic Data sets

## 5.3 GA parameters

Table 2 gives the parameters of GA

Table 2 *Parameters of GA*

| S.No. | Parameter | Initial value |
|-------|-----------|---------------|
| 1 | Maximum number of generations (MAXGEN) | 100 |
| 2 | Number of iterations (MAXITER) | 20 |
| 3 | Initial population size (POPSIZE) | 90 to 150 |
| 4 | Crossover Rate (CR) | 0.7 |
| 5 | Mutation probability (MU) | 0.05 |

## 5.4 Comparison Based on Mean Values

Table 3 Mean values of Vigilance and Tolerance

| Sl. No. | Data set | Algorithm | Avg. No. of clusters | Mean Error rate (%) | Mean Vig (ρ) | Mean Tolerances (δ) |
|---------|----------|-----------|---------------------|--------------------|--------------|--------------------|
| 1 | Iris | EK-FLANN | 3 | 7.88 | 0.7275 | 0.9396, 0.789, 1.6893, 0.7555 |
| | | K-FLANN | 3 | 8.4467 | 0.88 | 1.0036, 0.6921, 1.5719, 0.6309 |
| 2 | Wine | EK-FLANN | 3 | 7.1067 | 0.5667 | 0.3034, 0.2681, 0.3395, 0.3011, 0.3289, 0.2880, 0.2943, 0.2728, 0.2650, 0.2686, 0.2959, 0.2699, 0.2932 |
| | | K-FLANN | 3 | 7.8371 | 0.5833 | 0.3309, 0.3093, 0.3480,0.3016, 0.3265, 0.2665, 0.2854, 0.3196, 0.2757, 0.2732, 0.2980, 0.2840 0.2960 |
| 3 | New Thyroid | EK-FLANN | 3.06 | 5.7302 | 0.7660 | 0.3101, 0.3198, 0.3082, 0.2682, 0.2624 |
| | | K-FLANN | 3.87 | 10.2187 | 0.80 | 0.3218, 0.3041, 0.3167, 0.2981, 0.2687 |
| 4 | Haberman | EK-FLANN | 2 | 25.2723 | 0.5056 | 0.1646, 0.0521, 0.2502 |
| | | K-FLANN | 2 | 25.5283 | 0.5833 | 0.1725, 0.0486, 0.2602 |
| 5 | Syndata1 | EK-FLANN | 2 | 0.0020 | 0.95 | 4.4120, 0.9502 |
| | | K-FLANN | 2 | 0 | 0.8833 | 4.3581, 0.8880 |
| 6 | Syndata2 | EK-FLANN | 2 | 1.8133 | 0.6833 | 2.8552, 0.7232 |
| | | K-FLANN | 2 | 1.8 | 0.7667 | 2.9827, 0.8161 |
| 7 | Syndata3 | EK-FLANN | 2 | 28.91 | 0.7833 | 2.0556, 0.9249 |
| | | K-FLANN | 2 | 29.20 | 0.9667 | 2.3307, 0.9131 |
| 8 | Syndata4 | EK-FLANN | 3 | 0 | 0.9050 | 3.0562, 2.8014, 2.8995, .1648, 2.8471, 3.1377, 2.7869, 3.0127 |
| | | K-FLANN | 3 | 0 | 0.8950 | 3.0871, 2.7163, 3.2196, 3.1862, 2.9551, 2.8555, 2.9325, 2.8625 |
| 9 | Syndata5 | EK-FLANN | 3 | 0 | 0.9208 | 3.0242, 3.0091, 3.0510,3.3551, 3.3968, 3.1237, 3.0677, 3.1608 |
| | | K-FLANN | 3 | 0 | 0.9250 | 2.9942, 3.0177, 3.1081,3.3291, 3.4032, 3.1767, 2.9780, 3.1347 |
| 10 | Syndata6 | EK-FLANN | 2.24 | 50.8971 | 0.6075 | 1.5687, 1.764, 1.6202, 1.8806, 1.7013, 2.0719, 1.9648, 1.9283 |
| | | K-FLANN | 2.96 | 51.5429 | 0.6975 | 1.6992, 1.6767, 1.5112, .9602, 1.8124, 2.1032, 1.8294, 1.9093 |

Table 3 shows the mean values of parameters for 100 iterations. Both EK-FLANN and K-FLANN works well for well separated & half separated data sets (synthetic datasets 1, 2, 4 and 5) and 100% accuracy is obtained. For all the data sets, the performance of EK-FLANN is high when compared with K-FLANN (mean error rate is low). Mean vigilance value for all the data sets is high (more number of features contributed in clustering) for original K-FLANN when compared to EK-FLANN. It means that more number of features matched out of maximum features for a data set for K-FLANN. Both EK-FLANN and K-FLANN are not efficient in the case of overlapped data sets (synthetic data set 3 and synthetic data set 4). For Haberman data set, the mean error rate is approximately equal to 25% i.e. 25% of the data is not classified properly.

## 5.5 Comparison Based on Maximum Vigilance

Table 4 GA Results based on Maximum Vigilance over 100 independent runs

| Sl. No | Data set | Algorithm | Max. Vig. (ρ) | Tolerance (δ) | Clusters | Error rate (%) |
|---|---|---|---|---|---|---|
| 1 | Iris | EK-FLANN | 1 | 1.5209, 0.3951,2.1188, 0.4701 | 3 | 6.6667 |
| | | K-FLANN | 1 | 1.1352, 0.6115,2.1028, 0.5122 | 3 | 10 |
| 2 | Wine | EK-FLANN | 0.6923 | 0.1509, 0.3979,0.3381, 0.3359, 0.2067, 0.3185,0.3261, 0.1320, 0.0705, 0.2903,0.2611, 0.4063, 0.3159 | 3 | 8.9888 |
| | | K-FLANN | 0.7692 | 0.4557, 0.4014,0.4728, 0.4488, 0.4115, 0.4492,0.4150, 0.4658, 0.3723, 0.4423,0.4926, 0.4442, 0.4604 | 3 | 7.8652 |
| 3 | New Thyroid | EK-FLANN | 1 | 0.4602, 0.4884,0.3401, 0.3669, 0.2998 | 3 | 5.1163 |
| | | K-FLANN | 1 | 0.4251, 0.3326,0.4865, 0.3178, 0.2564 | 3 | 10.2326 |
| 4 | Haberman | EK-FLANN | 1 | 0.2721, 0.0569,0.4819 | 2 | 24.1830 |
| | | K-FLANN | 0.6667 | 0.0511, 0.0326,0.2625 | 2 | 25.4902 |
| 5 | Syndata1 | EK-FLANN | 1 | 6.3130, 1.3827 | 2 | 0 |
| | | K-FLANN | 1 | 6.2621, 1.4329 | 2 | 0 |
| 6 | Syndata2 | EK-FLANN | 1 | 4.4099, 1.3039 | 2 | 1.7 |
| | | K-FLANN | 1 | 1.5390, 1.28 | 2 | 1.8 |
| 7 | Syndata3 | EK-FLANN | 1 | 3.0764, 0.9713 | 2 | 28.8 |
| | | K-FLANN | 1 | 3.0764, 0.9713 | 2 | 29.20 |
| 8 | Syndata4 | EK-FLANN | 1 | 3.1151, 2.7980,3.4499, 3.2966, 3.6708, 2.2285,1.9557, 3.4605, | 3 | 0 |
| | | K-FLANN | 1 | 3.0849, 3.4785,2.6878, 3.0461, 3.0756, 2.6384,3.4341, 3.3365 | 3 | 0 |
| 9 | Syndata5 | EK-FLANN | 1 | 4.5344, 2.9597,2.5971, 3.6378, 4.0064, 2.8393,4.7230, 4.1131 | 3 | 0 |
| | | K-FLANN | 1 | 4.5344, 2.9597,2.5971, 3.6378, 4.0064, 2.8393,4.7230, 4.1131 | 3 | 0 |
| 10 | Syndata6 | EK-FLANN | 0.7500 | 1.6603, 2.8377,0.5326, 2.9208, 1.7214, 3.0582,2.9755, 1.1914 | 2 | 51.1429 |
| | | K-FLANN | 0.8750 | 1.5008, 2.6365,1.8204, 1.8916, 2.3631, 2.8480,1.4018, 2.8231 | 4 | 53.7143 |

In Table 4, the results are compared based on maximum vigilance over 100 generations. The results indicate that the EK-FLANN outperforms when compared with original K-FLANN.

# 6. CONCLUSION AND FUTURE WORK

In this work, the optimal values for vigilance and tolerance are determined using GA. and two K-FLANN models are compared. The above results show that EK-FLANN gives the accuracy about 90% for well separated and partially separated data sets. But the performance needs to be improved for highly overlapped data sets. So the genetic algorithm is efficient for finding the optimal values of vigilance ($\rho$) and tolerance ($\delta$). It is found that GA needs more execution time for large data sets. So, to improve the speed of the algorithm for large data sets efficient optimization technique is required..